\pdfoutput=1

\documentclass[11pt]{article}

\usepackage[preprint]{acl}

\usepackage{times}
\usepackage{latexsym}
\usepackage{subcaption}
\usepackage{booktabs}
\usepackage{amsmath,bm}

\usepackage[T1]{fontenc}

\usepackage[utf8]{inputenc}

\usepackage{microtype}

\usepackage{inconsolata}

\usepackage{graphicx}

\usepackage{dirtytalk}
\usepackage{enumitem}

\usepackage{etoolbox}
\patchcmd{\quote}{\rightmargin}{\leftmargin 1.8em \rightmargin}{}{}

\title{With Privacy, Size Matters: \\ On the Importance of Dataset Size in Differentially Private Text Rewriting}

\author{Stephen Meisenbacher \and Florian Matthes \\
Technical University of Munich\\
School of Computation, Information and Technology \\
Department of Computer Science\\
Garching, Germany\\
\texttt{\{stephen.meisenbacher,matthes\}@tum.de} \\
}

\begin{document}
\maketitle
\begin{abstract}
Recent work in Differential Privacy with Natural Language Processing (DP NLP) has proposed numerous promising techniques in the form of \textit{text rewriting} mechanisms. In the evaluation of these mechanisms, an often-ignored aspect is that of \textit{dataset size}, or rather, the effect of dataset size on a mechanism's efficacy for utility and privacy preservation. In this work, we are the first to introduce this factor in the evaluation of DP text privatization, where we design utility and privacy tests on large-scale datasets with dynamic split sizes. We run these tests on datasets of varying size with up to one million texts, and we focus on quantifying the effect of increasing dataset size on the privacy-utility trade-off. Our findings reveal that dataset size plays an integral part in evaluating DP text rewriting mechanisms; additionally, these findings call for more rigorous evaluation procedures in DP NLP, as well as shed light on the future of DP NLP in practice and at scale.
\end{abstract}

\section{Introduction}
In the study of Differential Privacy in Natural Language Processing (DP NLP), a multitude of approaches have been proposed, ranging from word-level obfuscation mechanisms to document-level rewriting techniques \cite{klymenko-etal-2022-differential,hu-etal-2024-differentially}.
To evaluate the efficacy of a proposed DP rewriting mechanism, researchers often perform privacy evaluations, where a defined attack vector is shown to be mitigated through the privatization of textual data \cite{feyisetan_balle_2020,mattern-etal-2022-limits,meisenbacher-etal-2024-comparative}. In addition, researchers may opt to demonstrate that a mechanism also preserves data utility, or usefulness in a downstream task \cite{10.1145/3485447.3512232, igamberdiev-etal-2022-dp}. These evaluations are based on a chosen set of datasets, which ideally simulate some plausibly \say{sensitive} scenario.

The primary factor in testing DP rewriting is the $\varepsilon$ parameter, also known as the \textit{privacy budget} in DP \cite{dwork2006differential}. Evaluating a given mechanism over a range of $\varepsilon$ values is useful in investigating its behavior over various privacy regimes, and it is generally true that lower $\varepsilon$ values lead to higher privacy and lower utility, and vice versa \cite{10.1145/2976749.2978318}. A secondary factor that is highly important, yet often not considered, is that of \textit{dataset size}, namely, how a mechanism behaves when run on datasets of varying size. While it is generally believed in DP that as dataset size grows, the effect of the injected random noise becomes less adverse \cite{ponomareva2023dp}, there is little empirical evidence of this in the DP NLP literature. Above all, previous experiments do not consider differently sized splits of a dataset, therefore only presenting a static view of the performance of DP rewriting mechanisms.

We address this gap by investigating the impact of dataset size on the privacy and utility preservation capabilities of DP text rewriting methods, quantifying the shift in the privacy-utility trade-off as dataset size grows. We are guided by the following research question: 

\begin{quote}
    \textit{What observations can be made regarding privacy and utility as dataset size grows in local DP text rewriting scenarios?}
\end{quote}

To answer this, we design a comprehensive evaluation consisting of four large-scale datasets, with which utility and privacy experiments are conducted using five differently sized subsamples of each dataset. By doing so, we empirically demonstrate the observable effects of DP text rewriting as dataset size grows. We critically analyze these results, which give way to an outlook on the benefits and challenges of DP text privatization at scale, a point which is often not addressed in the literature.

Our findings reveal that dataset size does indeed matter with DP text rewriting, as mechanisms behave differently at various dataset splits and privacy budgets. In addition, we learn that as dataset size increases, trade-offs from privatization generally become more favorable, making a case for DP text privatization at scale. Concretely, we make the following contributions to the field of DP NLP:

\begin{enumerate}
    \itemsep 0em
    \item We are the first to conduct experiments in DP text rewriting in a \textit{variable dataset size} setting.
    \item We empirically demonstrate and highlight the importance of dataset size in DP rewriting.
    \item We provide an outlook on future DP NLP work, focused on making it practical at scale.
\end{enumerate}

\section{Related Work}
Work in DP NLP primarily investigates addressing the challenges of meaningful and effective text privatization while also finding a balance with preserved utility \cite{10.1145/3485447.3512232}. Several recent works propose novel DP mechanisms for NLP \cite{hu-etal-2024-differentially}, often leveraging the \textit{local} model of DP \cite{4690986} for \textit{text rewriting}, and others have focused on investigating the challenges of DP NLP \cite{feyisetan2021research,mattern-etal-2022-limits,klymenko-etal-2022-differential,meisenbacher-etal-2025-impact}, as well as highlighting important points for moving the field forward \cite{meisenbacher-matthes-2024-thinking,vu-etal-2024-granularity}. In particular, the evaluation and benchmarking of DP NLP is seen as crucial yet challenging \cite{igamberdiev-etal-2022-dp,meisenbacher-etal-2024-comparative,arnold-2025-inspecting}, and recent works have proposed techniques for benchmarking DP text rewriting.

In previous works, evaluations often use publicly available datasets to mirror sensitive data scenarios. For example, datasets such as IMDb \cite{maas-etal-2011-learning}, Trustpilot \cite{10.1145/2736277.2741141}, or Yelp \cite{10.5555/2969239.2969312}, which contain some sensitive attributes (e.g., author ID or gender), are frequent choices for empirical privacy experiments. For utility evaluations, the choice of datasets is often varied \cite{meisenbacher-etal-2024-comparative}. In many recent local DP text rewriting works, the size of the evaluation datasets is quite small, particularly in privacy evaluations \cite{mattern-etal-2022-limits,Meehan2022SentencelevelPF, utpala-etal-2023-locally}. One notable exception is \textsc{DP-BART} \cite{igamberdiev-habernal-2023-dp}, which uses extensive datasets, yet without empirical privacy evaluations. Furthermore, in these works, the datasets are fixed in size, and the effect of privatization is not tested for different splits.

We view this to be a considerable gap, namely that DP text rewriting evaluations predominantly (1) do not generally test on larger-scale datasets, thus leaving it unclear whether proposed methods are effective at scale, and (2) do not vary the size of a given evaluation dataset, thus leaving the impact of dataset size on DP text rewriting unexplored. As such, we address these two shortcomings in this work, with the goal of showing the merit of extended evaluation setups in future DP NLP works.

\section{Experimental Setup and Results}
To investigate the effect of dataset size in DP text rewriting, we design utility and privacy experiments where performance is measured on large datasets with various split sizes, described next.

\subsection{Datasets and Tasks}
In our initial experiments, we utilize four datasets with associated utility and privacy tasks.

\paragraph{AG News.}
The \textit{AG News} dataset is a corpus prepared by \citet{10.5555/2969239.2969312} from a larger corpus of more than one million news articles. This subset contains 120k articles from four news domains, with 30k samples per class. The articles are shorter in nature, with an average word length of 43.95. We use this large-scale dataset to test the effect of DP text rewriting in a four-class classification task. 

\paragraph{MNLI.}
The \textit{Multi-Genre Natural Language Inference Corpus} (MNLI) is a dataset part of the GLUE benchmark \cite{wang-etal-2018-glue}, where MNLI contains \textit{premise} and \textit{hypothesis} pairs. The corresponding task is to classify whether the premise text \textit{entails}, \textit{contradicts}, or is \textit{neutral} to the hypothesis text. We utilize the entire dataset of 392,702 samples, with an average premise length of 22.28 words. For DP rewriting, we only privatize the premise text, leaving the hypothesis intact.

\paragraph{Trustpilot Reviews.}
The \textit{Trustpilot} corpus is a large-scale collection of user reviews. The corpus prepared by \citet{10.1145/2736277.2741141} tags each review with the stars provided (1-5), as well as the gender of the review writer. We utilize the \textit{en-US} split, taking all reviews with gender information (male / female), and filtering by negative reviews (1-2 stars) and positive reviews (5 stars). This results in a dataset of 366,210 reviews with an average of 52.39 words per review. We use this dataset for a utility task (binary sentiment analysis), as well as an adversarial privacy task (gender inference). 

\paragraph{Yelp Reviews.}
The \textit{Yelp Open Dataset} is a massive corpus of nearly seven million user reviews from Yelp. We take a smaller subset of the top-50 most frequently occurring users in the dataset\footnote{\scriptsize\url{https://business.yelp.com/data/resources/open-dataset/}}, which results in a dataset of 22,043 reviews with an average length of 196.29 words. This was done to facilitate a reasonable adversarial privacy task for authorship identification, which is paired with a utility task of binary sentiment analysis.

\subsection{DP Rewriting Methods}
For private rewriting of the chosen datasets, we choose three local DP rewriting mechanisms from the recent literature, which operate on three different levels of the syntactic hierarchy, i.e., yielding word-, token-, and document-level DP.

\paragraph{\textsc{1-Diffractor} \cite{10.1145/3643651.3659896}.} The \textsc{1-Diffractor} mechanism leverages metric local DP (MLDP) to obfuscate texts word-by-word in an efficient manner. We utilize the geometric version of the mechanism to rewrite texts with a word-level privacy budget of $\varepsilon \in \{0.5, 1, 3\}$, following values chosen in the original work.

\paragraph{\textsc{DP-Prompt} \cite{utpala-etal-2023-locally}.} The \textsc{DP-Prompt} mechanism uses a temperature sampling mechanism during token sampling in text generation. We choose the \textsc{flan-t5-large} model \cite{chung} as the underlying LM, and clip the logits to the range of (-95, 8)\footnote{\scriptsize Based on an empirical measurement of 100 randomly selected dataset texts, taking ($logit\_mean$, $logit\_mean + 4\cdot logit\_std$).}, which are then normalized to the range of [0, 1]. Choosing temperature values of $T \in \{1.75, 1.5, 1.25\}$, thus resulting in \textit{per-token} privacy budgets of $\varepsilon \in \{1.14, 1.\overline{3}, 1.6\}$\footnote{\scriptsize Following $\varepsilon = \frac{2\Delta}{T}$, where the sensitivity ($\Delta$) here is 1.}, respectively.

\paragraph{\textsc{DP-BART} \cite{igamberdiev-habernal-2023-dp}.} The \textsc{DP-BART} mechanism rewrites texts on the \textit{document-level} by adding DP noise to encoder representations in a BART model \cite{lewis-etal-2020-bart}. We use the base version of the mechanism (\textbf{DP-BART-CLV}) with a \textsc{bart-base} model and clipping range (-0.1, 0.1). For document-level privatization, we choose the budgets of $\varepsilon \in \{500, 1000, 1500\}$, following the original work. 

\paragraph{A note on comparability.}
We caution that the main focus of our experiments is to measure the effect of dataset size on privacy and utility metrics, and not to draw conclusions on the comparative effectiveness of the selected DP mechanisms. The latter would require a careful selection of privacy budget parameters to ensure proper comparability between the three mechanisms operating on different linguistic levels and therefore offering differing privacy guarantees. Instead, we follow the $\varepsilon$ values chosen by the original works, allowing for an analysis on the observable effects \textit{within} a mechanism across dataset sizes and privacy budgets.

\begin{table*}[htbp]
\centering
\resizebox{\linewidth}{!}{
\begin{tabular}{c|cccccccccc||cccccccccc|}
\multicolumn{1}{l|}{} & \multicolumn{10}{c||}{AG News} & \multicolumn{10}{c|}{MNLI} \\ \cline{2-21} 
\multicolumn{1}{l|}{} & \multicolumn{1}{l|}{} & \multicolumn{3}{c|}{\textsc{1-Diffractor}} & \multicolumn{3}{c|}{\textsc{DP-Prompt}} & \multicolumn{3}{c||}{\textsc{DP-BART}} & \multicolumn{1}{l|}{} & \multicolumn{3}{c|}{\textsc{1-Diffractor}} & \multicolumn{3}{c|}{\textsc{DP-Prompt}} & \multicolumn{3}{c|}{\textsc{DP-BART}} \\ \hline
\multicolumn{1}{l|}{Split \%} & \multicolumn{1}{c|}{Baseline} & 0.5 & 1 & \multicolumn{1}{c|}{3} & 1.14 & $1.\overline{3}$ & \multicolumn{1}{c|}{1.6} & 500 & 1000 & 1500 & \multicolumn{1}{c|}{Baseline} & 0.5 & 1 & \multicolumn{1}{c|}{3} & 1.14 & $1.\overline{3}$ & \multicolumn{1}{c|}{1.6} & 500 & 1000 & 1500 \\ \hline
0.1 & \multicolumn{1}{c|}{$92.3$} & \textit{88.8} & \textit{90.3} & \multicolumn{1}{c|}{$91.1$} & \textit{91.2} & $91.0$ & \multicolumn{1}{c|}{$91.0$} & \textit{53.0} & $81.0$ & \textit{87.1} & \multicolumn{1}{c|}{$87.1$} & $75.9$ & $80.1$ & \multicolumn{1}{c|}{\textit{85.0}} & $81.7$ & $81.9$ & \multicolumn{1}{c|}{$82.4$} & $60.9$ & \textit{67.6} & $69.4$ \\
0.25 & \multicolumn{1}{c|}{$92.4$} & $88.5$ & $89.8$ & \multicolumn{1}{c|}{$91.2$} & $90.3$ & $90.6$ & \multicolumn{1}{c|}{$90.8$} & \textit{55.6} & $81.4$ & $86.0$ & \multicolumn{1}{c|}{$88.0$} & $79.1$ & $82.0$ & \multicolumn{1}{c|}{$86.7$} & $83.4$ & $83.6$ & \multicolumn{1}{c|}{$84.0$} & \textit{63.2} & $70.0$ & $71.4$ \\
0.5 & \multicolumn{1}{c|}{$93.6$} & $90.2$ & $91.3$ & \multicolumn{1}{c|}{$92.9$} & $91.3$ & $91.5$ & \multicolumn{1}{c|}{$91.2$} & \textit{56.8} & $83.0$ & $86.6$ & \multicolumn{1}{c|}{$89.0$} & $80.0$ & $82.9$ & \multicolumn{1}{c|}{$87.5$} & $84.3$ & $84.4$ & \multicolumn{1}{c|}{$84.7$} & $64.9$ & $71.3$ & $73.1$ \\
0.75 & \multicolumn{1}{c|}{$93.9$} & $91.5$ & $92.4$ & \multicolumn{1}{c|}{$93.2$} & $91.9$ & $92.1$ & \multicolumn{1}{c|}{$92.2$} & \textit{57.0} & $83.9$ & $87.6$ & \multicolumn{1}{c|}{$89.4$} & $81.2$ & $83.8$ & \multicolumn{1}{c|}{$87.6$} & $84.6$ & $85.0$ & \multicolumn{1}{c|}{$85.2$} & $66.5$ & $72.6$ & $74.1$ \\
1 & \multicolumn{1}{c|}{$94.5$} & $91.8$ & $92.8$ & \multicolumn{1}{c|}{$93.9$} & $92.3$ & $92.5$ & \multicolumn{1}{c|}{$92.6$} & $57.7$ & $84.1$ & $87.6$ & \multicolumn{1}{c|}{$89.4$} & $81.6$ & $83.9$ & \multicolumn{1}{c|}{$87.8$} & $84.9$ & $85.2$ & \multicolumn{1}{c|}{$85.3$} & \textit{66.4} & $72.9$ & $74.4$
\end{tabular}
}
\caption{Utility Results for \textit{AG News} and \textit{MNLI}. The scores represent the average micro-F1 score of the respective experiment over three runs. For readability, the standard deviations are not reported; however, any average with a standard deviation in the range (0.3, 0.8] is \textit{italicized} (the highest recorded deviation was 0.8).}
\label{tab:utility}
\end{table*}

\begin{table*}[ht!]
\centering
    \resizebox{\textwidth}{!}{
    \begin{tabular}{cc|c|ccc|ccc|ccc||c|ccc|ccc|ccc}
    \multicolumn{2}{l}{} & \multicolumn{10}{|c||}{Trustpilot} & \multicolumn{10}{c}{Yelp} \\ \cline{3-22} 
& &  & \multicolumn{3}{c|}{\textsc{1-Diffractor}} & \multicolumn{3}{c|}{\textsc{DP-Prompt}} & \multicolumn{3}{c||}{\textsc{DP-BART}} & &  \multicolumn{3}{c|} {\textsc{1-Diffractor}} & \multicolumn{3}{c|}{\textsc{DP-Prompt}} & \multicolumn{3}{c}{\textsc{DP-BART}}\\ \cline{1-22} 
 \multicolumn{1}{c|}{Split \%} & \multicolumn{1}{c|}{} & Baseline & 0.5 & 1 & 3 & 1.14 & $1.\overline{3}$ & 1.6 & 500 & 1000 & 1500 & Baseline & 0.5 & 1 & 3 & 1.14 & $1.\overline{3}$ & 1.6 & 500 & 1000 & 1500\\ \hline
\multicolumn{1}{c|}{} & \multicolumn{1}{c|}{Util $\uparrow$} & 99.3 & \textit{97.0} & \textit{98.4} & \textit{98.5} & 98.1 & 98.5 & 98.5 & 94.0 & 97.8 & 98.6 & \textit{97.4} & \textit{94.6} & \textit{94.7} & \textit{96.8} & \textit{85.1} & 89.4 & \textit{91.7} & 82.1 & \textit{89.1} & 91.6 \\
\multicolumn{1}{c|}{} & \multicolumn{1}{c|}{Priv (s) $\downarrow$} & 72.8 & 65.1 & 69.1 & 72.1 & 69.5 & 68.8 & 69.5 & 59.1 & 60.7 & 62.4 & 8.1 & 7.7 & 7.7 & 7.2 & 4.1 & 3.2 & 3.6 & 6.3 & 7.7 & 6.8 \\
\multicolumn{1}{c|}{0.1} & \multicolumn{1}{c|}{Priv (a) $\downarrow$} & 72.8 & \textit{64.5} & \textit{62.4} & \textit{67.9} & \textit{68.1} & 68.3 & \textit{67.3} & 59.5 & \textit{62.0} & \textit{64.5} & 8.1 & \textit{5.0} & \textit{6.2} & \textit{6.6} & \textit{3.5} & \textit{3.6} & \textit{4.4} & \textit{4.7} & \textit{5.7} & \textit{7.7} \\
\multicolumn{1}{c|}{} & \multicolumn{1}{c|}{$\gamma$  (s) $\uparrow$} & - & -0.68 & -0.26 & -0.26 & -0.36 & \textbf{0.28} & \textbf{-0.22} & -1.61 & -0.34 & \textbf{-0.10} & - & -0.33 & -0.32 & 0.03 & -1.18 & -0.26 & -0.22 & -1.86 & -1.08 & -0.63 \\
\multicolumn{1}{c|}{} & \multicolumn{1}{c|}{$\gamma$  (a) $\uparrow$} & - & -0.67 & -0.16 & -0.20 & -0.34 & -0.21 & \textbf{-0.20} & -1.62 & -0.36 & \textbf{-0.12} & - & 0.00 & -0.13 & 0.10 & -1.11 & -0.53 & -0.32 & -1.66 & -0.83 & -0.74\\
\multicolumn{1}{c|}{} & \multicolumn{1}{c|}{NN $\uparrow$} & - & 204 & 33 & 1 & 521 & 474 & 401 & 5229 & 1068 & 520 & - & 1 & 1 & 1 & 53 & 59 & 49 & 590 & 121 & 55\\ \hline
\multicolumn{1}{c|}{} & \multicolumn{1}{c|}{Util $\uparrow$} & 99.7 & 98.3 & 99.1 & 99.6 & \textit{98.4} & 96.8 & 98.7 & 94.9 & \textit{96.5} & 98.8 & 97.5 & 92.8 & 94.2 & 96.7 & \textit{90.2} & \textit{93.4} & \textit{90.8} & \textit{80.3} & 90.8 & 93.1 \\
\multicolumn{1}{c|}{} & \multicolumn{1}{c|}{Priv (s) $\downarrow$} & 73.9 & 67.2 & 70.0 & 73.5 & 69.7 & 69.5 & 70.6 & 61.0 & 62.1 & 62.5 & 22.5 & 12.7 & 14.5 & 18.5 & 8.5 & 7.1 & 7.6 & 6.9 & 5.8 & 8.2 \\
\multicolumn{1}{c|}{0.25} & \multicolumn{1}{c|}{Priv (a) $\downarrow$} & 73.9 & \textit{65.5} & \textit{67.8} & \textit{73.6} & \textit{70.1} & 70.6 & \textit{69.4} & \textit{60.0} & \textit{63.2} & \textit{63.6} & 22.5 & 6.5 & \textit{7.0} & \textit{11.1} & \textit{6.2} & \textit{5.9} & \textit{6.4} & \textit{4.7} & \textit{5.7} & \textit{6.3}\\
\multicolumn{1}{c|}{} & \multicolumn{1}{c|}{$\gamma$  (s) $\uparrow$} & - & \textbf{-0.33} & -0.13 & -0.02 & -0.33 & -0.73 & -0.26 & \textbf{-1.27} & -0.80 & -0.12 & - & -0.09 & -0.02 & 0.09 & -0.20 & \textbf{2.68} & -0.09 & -1.24 & -0.01 & 0.14\\
\multicolumn{1}{c|}{} & \multicolumn{1}{c|}{$\gamma$  (a) $\uparrow$} & - & \textbf{-0.31} & -0.10 & -0.03 & -0.34 & -0.83 & -0.24 & \textbf{-1.26} & -0.82 & -0.13 & - & 0.18 & 0.32 & 0.42 & -0.10 & \textbf{0.28} & -0.04 & -1.14 & -0.01 & 0.23 \\
\multicolumn{1}{c|}{} & \multicolumn{1}{c|}{NN $\uparrow$} & - & 347 & 58 & 2 & 744 & 658 & 577 & 6546 & 1662 & 855 & - & 2 & 1 & 1 & 147 & 138 & 129 & 1446 & 316 & 130\\ \hline
\multicolumn{1}{c|}{} & \multicolumn{1}{c|}{Util $\uparrow$} & 99.7 & 98.0 & \textit{97.1} & 99.6 & 98.7 & 98.7 & 98.8 & \textit{94.2} & 98.3 & 98.7 & 98.8 & \textit{95.8} & \textit{97.7} & 98.6 & \textit{92.5} & 92.3 & \textit{92.2} & \textit{81.3} & \textit{92.1} & 94.1 \\
\multicolumn{1}{c|}{} & \multicolumn{1}{c|}{Priv (s) $\downarrow$} & 74.7 & 66.9 & 69.7 & 73.5 & 70.6 & 71.2 & 71.1 & 60.9 & 60.7 & 60.9 & 45.6 & 14.3 & 16.7 & 29.6 & 14.1 & 15.5 & 15.7 & 7.3 & 9.6 & 8.2 \\
\multicolumn{1}{c|}{0.5} & \multicolumn{1}{c|}{Priv (a) $\downarrow$} & 74.7 & \textit{63.7} & \textit{68.3} & \textit{73.3} & \textit{63.4} & \textit{65.7} & \textit{70.0} & 60.4 & 60.1 & \textit{63.4} & 45.6 & \textit{14.7} & \textit{19.2} & \textit{20.8} & \textit{8.1} & \textit{9.2} & \textit{8.9} & \textit{6.3} & \textit{9.4} & \textit{12.6}\\
\multicolumn{1}{c|}{} & \multicolumn{1}{c|}{$\gamma$  (s) $\uparrow$} & - & -0.42 & -0.74 & \textbf{-0.01} & \textbf{-0.25} & 0.05 & -0.23 & -1.52 & -0.25 & -0.12 & - & 0.40 & 0.53 & 0.33 & 0.09 & 0.61 & 0.03 & -0.82 & 0.15 & 0.37\\
\multicolumn{1}{c|}{} & \multicolumn{1}{c|}{$\gamma$  (a) $\uparrow$} & - & -0.38 & -0.72 & -0.01 & \textbf{-0.16} & \textbf{-0.19} & -0.22 & -1.51 & -0.24 & -0.16 & - & \textbf{0.39} & \textbf{0.47} & \textbf{0.52} & \textbf{0.22} & 0.18 & \textbf{0.18} & -0.80 & 0.16 & 0.28\\
\multicolumn{1}{c|}{} & \multicolumn{1}{c|}{NN $\uparrow$} & - & 478 & 84 & 2 & 893 & 801 & 710 & 7309 & 2122 & 1156 & - & 3 & 1 & 1 & 309 & 270 & 257 & 2916 & 621 & 272 \\ \hline
\multicolumn{1}{c|}{} & \multicolumn{1}{c|}{Util $\uparrow$} & 99.7 & \textit{98.2} & \textit{99.0} & 98.8 & 98.7 & 98.7 & 98.8 & 93.4 & 98.3 & 98.8 & 98.4 & \textit{96.5} & 97.3 & 98.4 & 92.9 & 93.2 & 92.6 & 83.7 & 92.7 & 95.2 \\
\multicolumn{1}{c|}{} & \multicolumn{1}{c|}{Priv (s) $\downarrow$} & 74.4 & 66.8 & 69.8 & 73.8 & 70.6 & 70.9 & 71.0 & 61.4 & 61.9 & 62.2 & 43.7 & 8.8 & 11.8 & 22.6 & 12.5 & 12.4 & 13.2 & 6.4 & 7.4 & 7.6 \\
\multicolumn{1}{c|}{0.75} & \multicolumn{1}{c|}{Priv (a) $\downarrow$} & 74.4 & \textit{63.8} & \textit{71.0} & \textit{69.6} & \textit{66.4} & \textit{69.2} & 71.9 & 59.8 & 59.8 & 66.3 & 43.7 & \textit{21.9} & \textit{27.1} & \textit{35.1} & 12.9 & \textit{15.9} & \textit{16.8} & 7.2 & \textit{8.3} & \textit{13.7} \\
\multicolumn{1}{c|}{} & \multicolumn{1}{c|}{$\gamma$  (s) $\uparrow$} & - & -0.37 & -0.16 & -0.28 & -0.27 & 0.05 & -0.24 & -1.82 & -0.28 & -0.12 & - & 0.60 & 0.62 & 0.48 & \textbf{0.15} & 0.79 & 0.10 & \textbf{-0.66} & \textbf{0.25} & \textbf{0.50}\\
\multicolumn{1}{c|}{} & \multicolumn{1}{c|}{$\gamma$  (a) $\uparrow$} & - & -0.33 & -0.18 & -0.22 & -0.21 & -0.25 & -0.25 & -1.80 & -0.25 & -0.18 & - & 0.30 & 0.27 & 0.20 & 0.14 & 0.10 & 0.02 & \textbf{-0.67} & \textbf{0.22} & \textbf{0.36}\\
\multicolumn{1}{c|}{} & \multicolumn{1}{c|}{NN $\uparrow$} & - & 572 & 106 & 3 & 983 & 881 & 791 & 7660 & 2405 & 1358 & - & 4 & 1 & 1 & 407 & 364 & 348 & 3989 & 911 & 398\\ \hline
\multicolumn{1}{c|}{} & \multicolumn{1}{c|}{Util $\uparrow$} & 99.7 & \textit{96.1} & 99.2 & 99.6 & 98.7 & 98.8 & 98.7 & 94.4 & 98.4 & 98.7 & 98.5 & 96.6 & 97.6 & 98.4 & 92.7 & \textit{92.6} & 93.1 & \textit{83.2} & 92.7 & 94.9 \\
\multicolumn{1}{c|}{} & \multicolumn{1}{c|}{Priv (s) $\downarrow$} & 75.2 & 67.0 & 70.3 & 74.0 & 71.0 & 71.1 & 71.1 & 61.1 & 62.0 & 62.5 & 63.5 & 12.0 & 14.6 & 24.5 & 21.0 & 21.5 & 21.3 & 7.2 & 10.7 & 11.5 \\
\multicolumn{1}{c|}{1} & \multicolumn{1}{c|}{Priv (a) $\downarrow$} & 75.2 & \textit{63.5} & \textit{63.7} & \textit{69.3} & \textit{69.4} & \textit{69.9} & \textit{70.4} & 60.1 & \textit{62.1} & \textit{70.4} & 63.5 & \textit{29.3} & \textit{44.9} & \textit{52.8} & \textit{19.6} & 20.7 & \textit{21.4} & 8.6 & \textit{17.1} & \textit{21.9}\\
\multicolumn{1}{c|}{} & \multicolumn{1}{c|}{$\gamma$  (s) $\uparrow$} & - & -0.99 & \textbf{-0.09} & \textbf{-0.01} & \textbf{-0.25} & 0.10 & -0.25 & -1.42 & \textbf{-0.22} & -0.14 & - & \textbf{0.62} & \textbf{0.68} & \textbf{0.60} & 0.08 & 0.64 & 0.11 & -0.67 & 0.24 & 0.45\\
\multicolumn{1}{c|}{} & \multicolumn{1}{c|}{$\gamma$  (a) $\uparrow$} & - & -0.94 & \textbf{0.00} & \textbf{0.05} & -0.23 & -0.20 & -0.24 & -1.41 & \textbf{-0.22} & -0.24 & - & 0.34 & 0.20 & 0.16 & 0.10 & 0.07 & \textbf{0.11} & -0.70 & 0.14 & 0.29\\
\multicolumn{1}{c|}{} & \multicolumn{1}{c|}{NN $\uparrow$} & - & 641 & 123 & 3 & 1045 & 937 & 845 & 7881 & 2607 & 1508 & - & 5 & 1 & 1 & 486 & 441 & 418 & 4695 & 1143 & 516
\end{tabular}
    }

\caption{Utility and Privacy Results for \textit{Trustpilot} and \textit{Yelp}. The best relative gains ($\gamma$) for each (\textit{mechanism}, $\varepsilon$) pair are \textbf{bolded}, for both the static (s) and adaptive (a) settings. \textit{Italicized} results denote standard deviations above 0.3.}
\label{tab:privacy}
\end{table*}

\subsection{Experimental Procedure}
First, each dataset is DP rewritten using the three chosen mechanisms and their privacy budgets, yielding nine private counterparts per dataset. These are then used to create five splits of various sizes: 10\%, 25\%, 50\%, 75\%, and 100\%. This results in a total of 45 private \say{datasets} per original dataset, or 180 datasets in total. 

Using these datasets, we first perform training and evaluation for the associated utility task for each dataset. To do this, we train a \textsc{deberta-v3-base} \cite{he2021deberta} model on a 90\% training split, and evaluate the trained model's performance on the 10\% val split. For utility tasks, we report the micro-F1 score from this evaluation.

The privacy experiments are conducted in a similar manner, where a \textsc{deberta-v3-base} model is now trained as an adversarial classification model (2-class gender inference or 50-class author identification). Following previous work, we perform the privacy tests in both the \textit{static} and \textit{adaptive} settings \cite{10.1145/3485447.3512232, mattern-etal-2022-limits, utpala-etal-2023-locally}, where the static adversary trains on the original (non-privatized) texts and is evaluated on the private val split, while the adaptive adversary trains on the private train split. 

In addition to reporting the micro-F1 of these evaluations, we also calculate the \textit{relative gain} ($\gamma$), which represents the trade-off between the paired utility and privacy tasks. For the calculation of $\gamma$), we define $P_o$, $U_o$ to represent the baseline (non-rewritten) privacy and utility scores, respectively, and $P_r$, $U_r$ be the scores observed on the privatized datasets. With this, relative gain is defined as $\gamma = (U_r / U_o) - (P_r / P_o)$, where higher scores are better. Note that we calculate the change in micro-F1 over majority-class guessing on the validation set for utility tasks, denoted $MG_u$ (utility), as the Trustpilot and Yelp datasets sentiment analysis tasks are imbalanced (positive \>\> negative); thus $RG = \frac{U_r - MG_u}{U_o - MG_u} - \frac{P_r}{P_o}$. The exact procedure for calculating $\gamma$ values is outlined in Appendix \ref{sec:appendix_rg}.

We also design a \textit{indistinguishability} test to measure the effect of dataset size on lending plausible deniability via nearest neighbors (\textit{NN}). For each private dataset, we iteratively swap a private text with its original counterpart, then run a nearest-neighbor search to measure at which $k$ the original text is the $k$-th nearest neighbor to the private text. A higher average $k$ value would thus imply that the private counterparts are relatively indistinguishable from the originals. For performance reasons, we limit the top-k search to 10,000; thus, a score of 10,000 would represent the best privacy.

\begin{figure}[ht]
    \centering
    \includegraphics[width=0.95\linewidth]{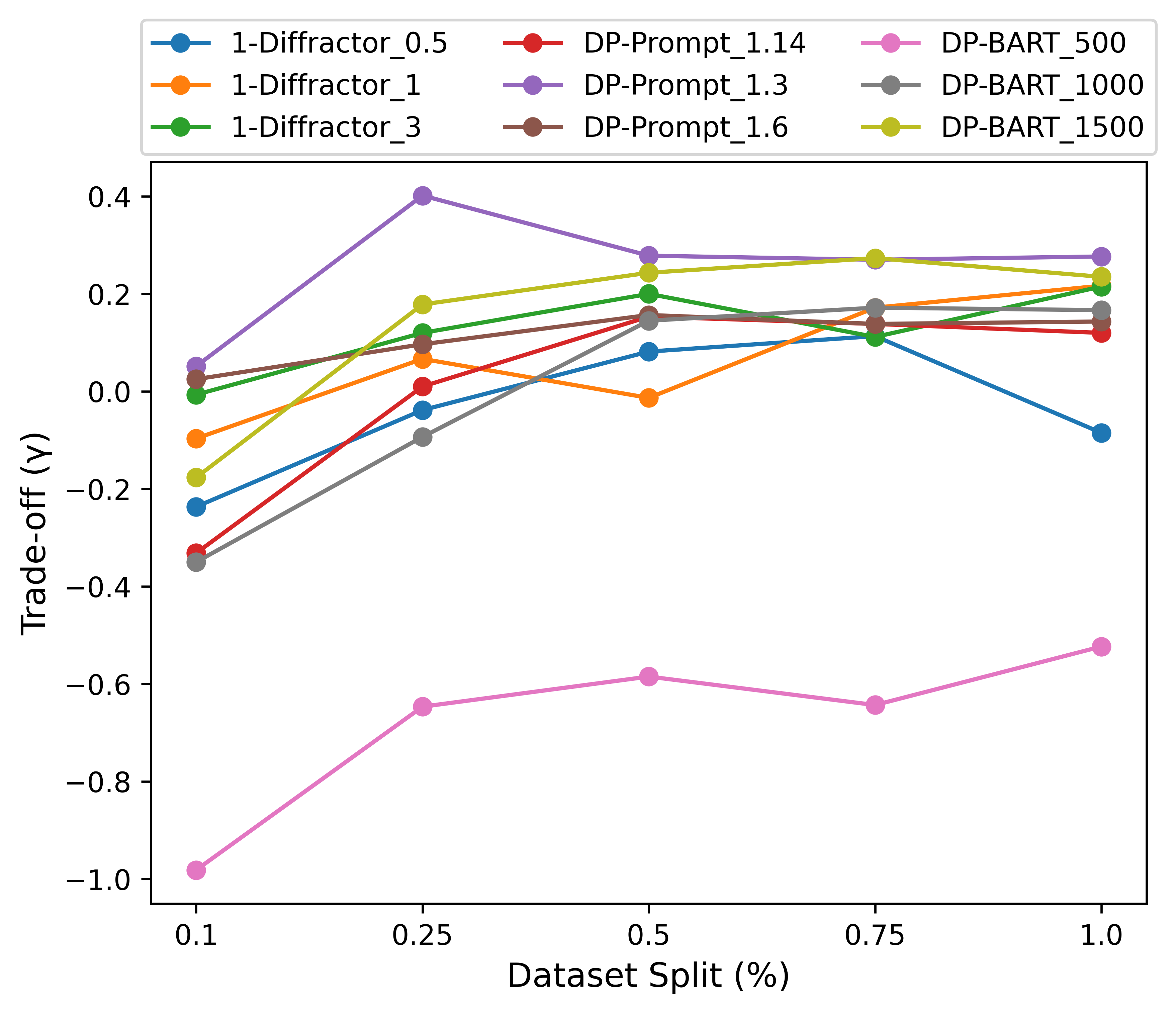}
    \vspace{-5pt}
    \caption{A visualization of average privacy-utility trade-offs ($\gamma$) per split percentage. For each mechanism-$\varepsilon$ pair, we average all trade-offs (both static and adaptive, over all three utilized privacy evaluation datasets).}
    \label{fig:to}
\end{figure}

All results are found in Table \ref{tab:utility} for AG News and MNLI, and Table \ref{tab:privacy} for Trustpilot and Yelp. We also plot the average trade-off ($\gamma$) value over all datasets, for each dataset split \% in Figure \ref{fig:to}.

\paragraph{Reproducibility.}
We note that for all experiments, a single-GPU server equipped with a 48GB Nvidia RTX A6000 GPU was utilized, along with 128GB of main memory. The entirety of the work required to complete the experiments described took approximately 2.5 months of GPU time on this system. All training procedures use HuggingFace Trainer default parameters, and they are repeated three times for each dataset on different shuffles of the train split. For all sampling procedures throughout this work, namely dataset split preparation and train/val splitting, a random seed of 42 was used.

\subsection{Scaling Up with Twitter}
Beyond the Trustpilot and Yelp datasets, we sought to validate our findings on a larger-scale dataset, as the largest size dataset (Trustpilot) is limited to 360k reviews. To the best of the authors' knowledge, there does not exist a publicly available dataset that allows for the two-sided experiments we conduct in this work, namely with an associated utility \textit{and} privacy task. Therefore, we constructed a new dataset based on an existing very large-scale corpus of Tweets \cite{enryu_2023_15086029}.

We decided to focus on the \textit{authorship identification} adversarial task, as with the Yelp dataset. To construct a fitting dataset from the Twitter corpus, which contains around 88 million tweets, we first iterated through the corpus and enumerated the number of tweets authored by each unique user (by ID). Before this, however, we removed tweets detected not to be in English\footnote{\scriptsize Using \textsc{lingua-py} (\url{https://github.com/pemistahl/lingua-py})}. Then, we kept only the tweets from the top 1000 users, each of whom has 1001 tweets attributed to them (the maximum of any author in the dataset). This resulted in a dataset of just over one million tweets, which we give the title \textit{Twitter\_1Kx1K}\footnote{\scriptsize The dataset is publicly available at \url{https://huggingface.co/datasets/sjmeis/Twitter_1Kx1K}}.

\begin{table*}[t!]
\centering
    \resizebox{\linewidth}{!}{
    \begin{tabular}{cc|c|ccc|ccc|ccc|}
    \multicolumn{2}{l}{} & \multicolumn{7}{|c|}{Twitter} \\ \cline{3-9} 
& &  & \multicolumn{3}{c|}{\textsc{1-Diffractor}} & \multicolumn{3}{c|}{\textsc{DP-Prompt}} \\ \cline{1-9} 
 \multicolumn{1}{c|}{Split \%} & \multicolumn{1}{c|}{} & Baseline & 0.5 & 1 & 3 & 1.14 & $1.\overline{3}$ & 1.6 \\ \hline
\multicolumn{1}{c|}{} & \multicolumn{1}{c|}{Util $\uparrow$} & 91.2 & 80.9 & 84.1 & 89.4 & 49.8 & 51.1 & 52.1 \\
\multicolumn{1}{c|}{} & \multicolumn{1}{c|}{Priv (s) $\downarrow$} & 2.8 & 2.0 & 2.1 & 2.5 & 0.2 & 0.1 & 0.5 \\
\multicolumn{1}{c|}{0.1} & \multicolumn{1}{c|}{Priv (a) $\downarrow$} & 2.8 & 2.3 & 2.3 & 2.3 & 0.1 & 0.1 & 0.2 \\
\multicolumn{1}{c|}{} & \multicolumn{1}{c|}{$\gamma$  (s) $\uparrow$} & - & 0.19 & 0.17 & 0.11 & 0.49 & \textbf{0.51} & 0.40 \\
\multicolumn{1}{c|}{} & \multicolumn{1}{c|}{$\gamma$  (a) $\uparrow$} & - & 0.07 & 0.12 & \textbf{0.18} & 0.51 & \textbf{0.52} & 0.51 \\
\multicolumn{1}{c|}{} & \multicolumn{1}{c|}{NN $\uparrow$} & - & 4983 & 4986 & 4991 & 4970 & 4992 & 5002 \\ \hline
\multicolumn{1}{c|}{} & \multicolumn{1}{c|}{Util $\uparrow$}  & 95.1 & 84.2 & 87.4 & 93.4 & 50.4 & 50.7 & 51.8 \\
\multicolumn{1}{c|}{} & \multicolumn{1}{c|}{Priv (s) $\downarrow$} & 10.6 & 6.6 & 7.4 & 8.3 & 0.2 & 0.4 & 1.0 \\
\multicolumn{1}{c|}{0.25} & \multicolumn{1}{c|}{Priv (a) $\downarrow$} & 10.6 & \textit{8.7} & \textit{8.6} & \textit{9.8} & 0.1 & 0.1 & 0.3 \\
\multicolumn{1}{c|}{} & \multicolumn{1}{c|}{$\gamma$  (s) $\uparrow$} & - & 0.26 & 0.22 & 0.20 & \textbf{0.51} & 0.49 & 0.45 \\
\multicolumn{1}{c|}{} & \multicolumn{1}{c|}{$\gamma$  (a) $\uparrow$} & - & 0.06 & 0.11 & 0.06 & \textbf{0.52} & \textbf{0.52} & 0.51 \\
\multicolumn{1}{c|}{} & \multicolumn{1}{c|}{NN $\uparrow$} & - & 7989 & 7988 & 8009 & 7968 & 8010 & 7994 \\ \hline
\multicolumn{1}{c|}{} & \multicolumn{1}{c|}{Util $\uparrow$} & 96.3 & 85.3 & 88.5 & 94.4 & 49.9 & 50.4 & 52.4 \\
\multicolumn{1}{c|}{} & \multicolumn{1}{c|}{Priv (s) $\downarrow$} & 26.8 & 15.4 & 16.5 & 18.4 & 0.3 & 0.5 & 1.4 \\
\multicolumn{1}{c|}{0.5} & \multicolumn{1}{c|}{Priv (a) $\downarrow$} & 26.8 & \textit{18.6} & \textit{22.6} & \textit{24.1} & 0.1 & 0.1 & 0.5 \\
\multicolumn{1}{c|}{} & \multicolumn{1}{c|}{$\gamma$  (s) $\uparrow$} & - & 0.31 & 0.30 & 0.29 & \textbf{0.51} & 0.50 & 0.49 \\
\multicolumn{1}{c|}{} & \multicolumn{1}{c|}{$\gamma$  (a) $\uparrow$} & - & \textbf{0.19} & 0.08 & 0.08 & 0.51 & \textbf{0.52} & \textbf{0.52} \\
\multicolumn{1}{c|}{} & \multicolumn{1}{c|}{NN $\uparrow$} & - & 8994 & 9000 & 9003 & 8987 & 9013 & 9003 \\ \hline
\multicolumn{1}{c|}{} & \multicolumn{1}{c|}{Util $\uparrow$} & 96.9 & 85.7 & 89.1 & 95.0 & 50.2 & 50.5 & \textit{52.2} \\
\multicolumn{1}{c|}{} & \multicolumn{1}{c|}{Priv (s) $\downarrow$} & 40.1 & 21.8 & 23.8 & 26.8 & 0.3 & 0.6 & 1.7 \\
\multicolumn{1}{c|}{0.75} & \multicolumn{1}{c|}{Priv (a) $\downarrow$} & 40.1 & \textit{29.8} & \textit{30.5} & \textit{32.2} & 0.1 & 0.1 & \textit{0.6} \\
\multicolumn{1}{c|}{} & \multicolumn{1}{c|}{$\gamma$  (s) $\uparrow$} & - & 0.34 & 0.32 & 0.31 & \textbf{0.51} & 0.50 & 0.49 \\
\multicolumn{1}{c|}{} & \multicolumn{1}{c|}{$\gamma$  (a) $\uparrow$} & - & 0.14 & \textbf{0.16} & \textbf{0.18} & 0.51 & 0.51 & \textbf{0.52} \\
\multicolumn{1}{c|}{} & \multicolumn{1}{c|}{NN $\uparrow$} & - & 9330 & 9334 & 9338 & 9322 & 9341 & 9332 \\ \hline
\multicolumn{1}{c|}{} & \multicolumn{1}{c|}{Util $\uparrow$} & 97.3 & 86.3 & 89.5 & 95.4 & 50.0 & \textit{50.6} & 52.3 \\
\multicolumn{1}{c|}{} & \multicolumn{1}{c|}{Priv (s) $\downarrow$} & 45.8 & 23.9 & 26.0 & 29.2 & 0.3 & 0.6 & 1.8 \\
\multicolumn{1}{c|}{1} & \multicolumn{1}{c|}{Priv (a) $\downarrow$} & 45.8 & \textit{36.0} & \textit{34.8} & \textit{38.0} & \textit{0.1} & \textit{0.1} & \textit{0.7} \\
\multicolumn{1}{c|}{} & \multicolumn{1}{c|}{$\gamma$  (s) $\uparrow$} & - & \textbf{0.36} & \textbf{0.35} & \textbf{0.34} & \textbf{0.51} & 0.50 & \textbf{0.50} \\
\multicolumn{1}{c|}{} & \multicolumn{1}{c|}{$\gamma$  (a) $\uparrow$} & - & 0.10 & \textbf{0.16} & 0.15 & 0.51 & 0.51 & \textbf{0.52} \\
\multicolumn{1}{c|}{} & \multicolumn{1}{c|}{NN $\uparrow$} & - & 9498 & 9502 & 9504 & 9495 & 9502 & 9502
\end{tabular}
\quad
    \begin{tabular}{cc|c|ccc|}
    \multicolumn{2}{l}{} & \multicolumn{4}{|c|}{Twitter} \\ \cline{3-6} 
& & & \multicolumn{3}{c|}{\textsc{DP-BART}} \\ \cline{1-6} 
 \multicolumn{1}{c|}{Split \%} & \multicolumn{1}{c|}{} & Baseline & 500 & 1000 & 1500 \\ \hline
\multicolumn{1}{c|}{} & \multicolumn{1}{c|}{Util $\uparrow$} & \textit{59.5} & 49.8 & 50.9 & 54.8 \\
\multicolumn{1}{c|}{} & \multicolumn{1}{c|}{Priv (s) $\downarrow$} & 2.8 & 1.1 & 1.1 & 2.0 \\
\multicolumn{1}{c|}{0.1} & \multicolumn{1}{c|}{Priv (a) $\downarrow$} & 2.8 & 1.2 & \textit{2.3} & 1.7 \\
\multicolumn{1}{c|}{} & \multicolumn{1}{c|}{$\gamma$  (s) $\uparrow$} & - & 0.44 & 0.46 & 0.20 \\
\multicolumn{1}{c|}{} & \multicolumn{1}{c|}{$\gamma$  (a) $\uparrow$} & - & 0.42 & 0.05 & 0.33 \\
\multicolumn{1}{c|}{} & \multicolumn{1}{c|}{NN $\uparrow$} & - & 0.42 & 0.05 & 0.33 \\ \hline
\multicolumn{1}{c|}{} & \multicolumn{1}{c|}{Util $\uparrow$}  &  \textit{77.0} & 54.2 & \textit{54.8} & 53.8\\
\multicolumn{1}{c|}{} & \multicolumn{1}{c|}{Priv (s) $\downarrow$} & 10.6 & 1.7 & 1.2 & 1.9 \\
\multicolumn{1}{c|}{0.25} & \multicolumn{1}{c|}{Priv (a) $\downarrow$} & 
10.6 & 2.2 & 2.3 & 2.9 \\
\multicolumn{1}{c|}{} & \multicolumn{1}{c|}{$\gamma$  (s) $\uparrow$} & - & 0.54 & \textbf{0.59} & 0.52 \\
\multicolumn{1}{c|}{} & \multicolumn{1}{c|}{$\gamma$  (a) $\uparrow$} &  - & 0.49 & 0.49 & 0.43 \\
\multicolumn{1}{c|}{} & \multicolumn{1}{c|}{NN $\uparrow$} & - & 7997 & 8012 & 8003 \\ \hline
\multicolumn{1}{c|}{} & \multicolumn{1}{c|}{Util $\uparrow$} & \textit{84.8} & 55.1 & 51.8 & 54.2 \\
\multicolumn{1}{c|}{} & \multicolumn{1}{c|}{Priv (s) $\downarrow$} & 26.8 & 1.5 & 1.5 & 1.5 \\
\multicolumn{1}{c|}{0.5} & \multicolumn{1}{c|}{Priv (a) $\downarrow$} & 26.8 & 2.7 & 2.9 & 3.5 \\
\multicolumn{1}{c|}{} & \multicolumn{1}{c|}{$\gamma$  (s) $\uparrow$} & - & \textbf{0.59} & 0.55 & \textbf{0.58} \\
\multicolumn{1}{c|}{} & \multicolumn{1}{c|}{$\gamma$  (a) $\uparrow$} & - & \textbf{0.55} & 0.50 & \textbf{0.51} \\
\multicolumn{1}{c|}{} & \multicolumn{1}{c|}{NN $\uparrow$} & - & 8986 & 9003 & 8993 \\ \hline
\multicolumn{1}{c|}{} & \multicolumn{1}{c|}{Util $\uparrow$} & \textit{88.7} & 54.3 & 53.5 & 53.3\\
\multicolumn{1}{c|}{} & \multicolumn{1}{c|}{Priv (s) $\downarrow$} & 40.1 & 1.5 & 1.3 & 1.2 \\
\multicolumn{1}{c|}{0.75} & \multicolumn{1}{c|}{Priv (a) $\downarrow$} & 40.1 & 3.5 & 3.4 & 3.5 \\
\multicolumn{1}{c|}{} & \multicolumn{1}{c|}{$\gamma$  (s) $\uparrow$} & - & 0.57 & 0.57 & 0.57\\
\multicolumn{1}{c|}{} & \multicolumn{1}{c|}{$\gamma$  (a) $\uparrow$} &  - & 0.52 & \textbf{0.52} & \textbf{0.51} \\
\multicolumn{1}{c|}{} & \multicolumn{1}{c|}{NN $\uparrow$} & - & 9326 & 9327 & 9333\\ \hline
\multicolumn{1}{c|}{} & \multicolumn{1}{c|}{Util $\uparrow$} & 90.6 & 53.5 & 53.6 & 53.4 \\
\multicolumn{1}{c|}{} & \multicolumn{1}{c|}{Priv (s) $\downarrow$} &  45.8 & 1.3 & 1.4 & 1.7 \\
\multicolumn{1}{c|}{1} & \multicolumn{1}{c|}{Priv (a) $\downarrow$} & 45.8 & 4.0 & 3.9 & 4.0 \\
\multicolumn{1}{c|}{} & \multicolumn{1}{c|}{$\gamma$  (s) $\uparrow$} & - & 0.56 & 0.56 & 0.55\\
\multicolumn{1}{c|}{} & \multicolumn{1}{c|}{$\gamma$  (a) $\uparrow$} & - & 0.50 & 0.50 & 0.50 \\
\multicolumn{1}{c|}{} & \multicolumn{1}{c|}{NN $\uparrow$} & - & 9493 & 9496 & 9501
\end{tabular}
    }
\caption{Utility and Privacy Results for \textit{Twitter}. On the left-hand side are presented the results from \textsc{1-Diffractor} and \textsc{DP-Prompt} on the \textit{Twitter\_1Kx1K} dataset, whereas the right-hand side presents the results from \textsc{DP-BART} on the smaller \textit{Twitter\_100x1K} subset. The best relative gains ($\gamma$) for each (\textit{mechanism}, $\varepsilon$) pair are \textbf{bolded}, for both the static (s) and adaptive (a) settings. \textit{Italicized} results denote standard deviations above 0.3.}
\label{tab:privacy_twitter}
\end{table*}

To connect this dataset with a utility task, we assign a sentiment score to each tweet. We use the simple Vader \textsc{SentimentIntensityAnalyzer} from \textsc{nltk} \cite{bird2009natural} to assign one of \textit{positive} (1), \textit{neutral} (0), or \textit{negative} (-1) to a tweet. We use the \textit{compound} score calculated by Vader, given on the scale of -1 to 1, and follow the guidelines of the original implementation\footnote{\scriptsize\url{https://github.com/cjhutto/vaderSentiment}}, namely scores above 0.5 as positive, below -0.5 as negative, and neutral for the rest. The resulting sentiments were distributed in roughly a 50/33/17 positive/neutral/negative split. Thus, we allow for a three-class sentiment classification task. The dataset is released with a \textsc{cc-by-4.0} license, mirroring the original corpus from \citet{enryu_2023_15086029}.

Following the dataset creation step, we followed the same privatization procedure with the Twitter data as was performed for Trustpilot and Yelp, using the three chosen DP mechanisms. Note that in the case of \textsc{DP-BART}, we limit the dataset to only the top 100 authors (i.e., a 10\% split), due to the considerable computation hours required to run \textsc{DP-BART}. We call this the \textit{Twitter\_100x1K} subset. Following privatization, we followed the same training and metric calculation procedures as with Trustpilot and Yelp. The results are found in Table \ref{tab:privacy_twitter}. Additionally, the detailed calculations for the $\gamma$ metric is included as part of Appendix \ref{sec:appendix_rg}.

\section{Regression Analysis}
Building on the results in Tables \ref{tab:privacy} and \ref{tab:privacy_twitter}, we conduct a regression analysis to quantify the effect of dataset size on the DP text rewriting, along with other associated variables. As a target variable, i.e., the dependent variable, we choose to predict relative gain ($\gamma$), as this gives an overall perspective on the effectiveness of DP text rewriting. In particular, we set the target to be the \textit{average} $\gamma$, namely the average between the static and adaptive $\gamma$ scores.

To predict the $\gamma$ score, we define a number of dependent variables, which are based on our experimental setup and results:
\begin{itemize}
    \itemsep 0em
    \item \textbf{Dataset size}: we use the size of the corresponding datasets in all of our experimental runs, considering the split size based on the \textit{split \%} of the overall dataset. Due to the very disparate range of dataset sizes, we take the natural logarithm of this value.
    \item \textbf{Average number of words}: we assume that an important quantity in the prediction of relative gains is not only the size of the dataset, but also the average text length within the dataset. As such, we calculate the average word length of each dataset split, using the \textsc{nltk.word\_tokenize} function.
    \item \textbf{Mechanism type}: we convert our three selected mechanisms into a categorical variable, specifically \textsc{1-Diffractor}=1, \textsc{DP-Prompt}=2, and \textsc{DP-BART}=3. While we realize that the intricacies of different DP mechanisms are much more complex, we justify this decision by considering the lexical level on which these mechanisms operate, i.e., ranging 1=\textit{word-level} to 3=\textit{document-level}.
    \item $\bm{\varepsilon}$: as with mechanism, we also convert the three tested $\varepsilon$ values per mechanism into a categorical (1, 2, 3) variable, as the precise values are mechanism-specific and not directly comparable. As such, we interpret this categorical variable as \textit{low}, \textit{medium}, and \textit{high} privacy budgets, or, from strictest to least strict privacy.
    \item \textbf{Utility labels}: this represents a simple count of the \textit{support} of the label set of the associated utility task, as we presume this has some impact on the relative gain.
    \item \textbf{Privacy labels}: similarly, we include the support of the \textit{privacy labels} for the associated (adversarial) privacy task. We take the natural logarithm of this value.
\end{itemize}

With these, we run a multivariate Ordinary Least Squares regression using the \textsc{statsmodels} library, with all default settings. Summary statistics of the fitted OLS model are provided in Table \ref{tab:mlr}.

\begin{table}[ht!]
\small
\centering
\resizebox{0.95\linewidth}{!}{
\begin{tabular}{r|llll}
\toprule
\multicolumn{1}{l}{$R^2=0.546$} & \multicolumn{1}{c}{\textbf{coef.}} & \multicolumn{1}{c}{\textbf{std err}} & \multicolumn{1}{c}{\textbf{t}} & \multicolumn{1}{c}{\textbf{P>|t|}} \\
\midrule
\midrule
constant & -8.2040 & 1.282 & -6.399 & 0.000 \\
size (log) & 0.1359	& 0.039 & 3.497 & 0.001 \\
avg. \# words & 0.0126 & 0.002 & 5.775 & 0.000 \\
mechanism & -0.1827 & 0.045 & -4.039 & 0.000 \\
$\varepsilon$ & 0.1743 & 0.039 & 4.449 & 0.000 \\
\# util. labels & 2.8812 & 0.454 & 6.350 & 0.000 \\
\# priv. labels (log) & -0.3209 & 0.072 & -4.465 & 0.000 \\
\bottomrule
\end{tabular}
}
\caption{MLR to predict the average $\gamma$ score. In general, $R^2$ measures the goodness of the fit, ranging from 0 to 1. All predictors are statistically significant.}
\label{tab:mlr}
\vspace{-10pt}
\end{table}

\subsection{Post-hoc Tests}
Following the OLS regression, we conduct a deeper analysis of the results, asking the question of whether there exists any significant differences \textit{between} dataset sizes, beyond the previous finding that there exists \textit{some} significant relationship between dataset size and relative gains.

To prepare this test, we first bin the (log) size variable into five equal bins calculated from the min and max observed values (using \textsc{pandas.cut}), resulting in the following bins:
\begin{center}
\footnotesize
    \{\textbf{1:} (7.698, 8.922] < \textbf{2:} (8.922, 10.145] < \textbf{3:} (10.145, 11.369] < \textbf{4:} (11.369, 12.593] < \textbf{5:} (12.593, 13.817]\}
\end{center}

We convert these bins into a categorical variable, namely 1 for the lowest bin and 5 for the highest. Following this, we run a Kruskal-Wallis analysis of variance (using \textsc{scipy.stats}), which results in a p-value of 0.00087, thus allowing us to reject the null hypothesis that the population medians of all size groups are equal. Since we do not make any assumption of normality, we then proceed to run a Dunn's post-hoc test \cite{dunn1964multiple}, which informs us of any significant differences between the categorical size variables. The results of this test are given in Table \ref{tab:dunn}, which finds several significant pairwise relations between binned dataset sizes.

\begin{table}[htbp]
    \centering
    \small
\resizebox{\linewidth}{!}{        
\begin{tabular}{l|ccccc}
 & \textbf{1} & \textbf{2} & \textbf{3} & \textbf{4} & \textbf{5}\\ \hline
\textbf{1} & 1.0000 & \textbf{0.0000} & 0.1338 & 0.3060 & \textbf{0.0148} \\
\textbf{2} & \textbf{0.0000} & 1.0000 & \textbf{0.0231} & \textbf{0.0001} & 0.0593 \\
\textbf{3} & 0.1338 & \textbf{0.0231} & 1.0000 & 0.4332 & 0.4987 \\
\textbf{4} & 0.3060 & \textbf{0.0001} & 0.4332 & 1.0000 & 0.0660 \\
\textbf{5} & \textbf{0.0148} & 0.0593 & 0.4988 & 0.0660 & 1.0000
\end{tabular}
}
\caption{Dunn's post-hoc test results. \textbf{Bolded} $p$-values indicate statistically significant results ($p < 0.05$).}
\label{tab:dunn}
\end{table}

\section{Discussion}
We reflect on the main findings of our experiments and discuss their implications for DP NLP.

\paragraph{Dataset size is important.}
Studying the results of our experiments, we find that dataset size is a very important factor when judging the efficacy of DP text rewriting, both from a utility and privacy standpoint. Looking to Table \ref{tab:privacy}, one can see that given a single mechanism, the \say{best-performing} configuration, i.e., in terms of relative gains, can vary depending on dataset size and task. As an example, in the Yelp task, \textsc{DP-BART} only begins to exhibit positive relative gains at larger dataset split sizes, and generally speaking, all three tested mechanisms only begin to demonstrate positive relative gains at or above the 50\% split mark. 

Looking to the regression results, we find that dataset size is indeed a significant factor in predicting the expected relative gain for DP text rewriting. In particular, our fitted model suggests a relatively strong positive relationship between dataset size and relative gain, namely that with higher dataset sizes, we can expect average relative gains to rise. The post-hoc tests provide a deeper view into this significant relationship, highlighting further evidence of differences in observed trade-offs when the size of the data to be privatized varies.

Beyond this, an important finding lies in the fact that given different privacy budgets with the same mechanism, the most favorable trade-offs may not always be with the same dataset size. This supports the idea that more dynamic testing in terms of dataset size should be performed to provide a more holistic evaluation picture. Such testing is especially needed considering the task-specific differences that may be present (e.g., Trustpilot vs. Yelp), where the factors of adversarial task, dataset size, and mechanism effects can all interplay in a complex way. This is discussed next.

\paragraph{What factors matter for favorable privacy-utility trade-offs?}
Our regression analysis sheds light on this matter, as well as highlights remaining gaps in understanding. Importantly, we find that all included independent variables in our analysis prove to be significant to some degree in predicting relative gains from DP text rewriting. This not only includes dataset size, as discussed above, but also the \textit{makeup} of the dataset, including average text length, the associated utility task, and even the nature of adversarial risk related to the dataset. 

Beyond this, we find that the choice of DP mechanism is also a significant factor, yet the coefficient of \textit{mechanism} implies that as we approach document-level methods (e.g., \textsc{DP-BART}), we might expect average relative gains to decrease. On the other hand, we observe from the empirical results that \textsc{DP-BART} on average experiences the largest \textit{absolute} improvements as dataset size increases; however this comes with the caveat of often starting at a much lower point. We therefore observe the effect of document-level methods, where significant noise leads to a poor starting point, but with more data, may still exhibit utility.  These are important findings, as the results suggest that mechanism specifics are significant in influencing expected trade-offs, providing motivation for novel methods that operate on diverse linguistic units. 

An interesting insight comes with the regression results related to $\varepsilon$, as the coefficient suggests that as $\varepsilon$ increases (i.e., the privacy guarantees become weaker), the average relative gains tend to increase. This may initially seem like a discouraging result for the field of DP text rewriting, yet one must consider that along this line, there exists some balance between chosen $\varepsilon$ and (reasonably) positive relative gain. Nevertheless, the true relationship between $\varepsilon$ and the privacy-utility trade-off most likely runs much deeper, and also is certainly intertwined with that of \textit{mechanism} and other variables.

Despite the rich insights provided by conducting a regression analysis on the experiment results, we learn that there still exist gaps in understanding what exactly influences the expected outcomes from DP text rewriting. This is signified by the $R^2$ score of our fitted model (0.546), which suggests a good fit, yet one with a relatively high amount of variance still unexplained. Therefore, this implies that there still exist potentially numerous other factors important to the picture, which we simply did not include as variables in fitting the model. Additionally, some correlations (such as increasing $\varepsilon$ leading to increasing $\gamma$) must be tested more rigorously at scale. Finally, the effect of \textit{task setup}, i.e., the number of utility and privacy \say{labels}, is only surfaced but merits further investigation. Such results provides motivation for future research into achieving a better understanding of the factors influencing local DP text rewriting outcomes.

\paragraph{There \textit{is} utility in DP.}
A promising result, which can be extracted from both Tables \ref{tab:utility} and \ref{tab:privacy}, is that as dataset size grows larger, downstream utility of DP privatized text also generally increases. While this would generally be assumed outside of the DP context, these findings show promise that there is indeed utility to be gained from privatized text, and that these gains can become more apparent with larger dataset sizes. In particular, Table \ref{tab:utility} demonstrates that as more data is available, DP text outpaces non-private text in \say{closing the gap} to the highest attainable utility. Albeit, this is due in part to the lower starting point of privatized data, yet the findings here show that using smaller dataset sizes may in fact be showing a \textit{lower bound} of downstream utility for DP rewritten text. 

\paragraph{DP rewriting at scale? An outlook.}
The above considerations lead to a very important question: can DP text rewriting, or more generally DP text privatization, make sense \textit{at scale}? Such a discussion is crucial in weighing the potential benefits of DP text privatization in practical settings.

In favor of the affirmative answer to this question are several findings from this work. Firstly, we can observe that (significant) positive relative gains are possible in certain scenarios, particularly in those with larger dataset sizes. Beyond this, the \textit{NN} metric clearly shows the \textit{indistinguishability in a crowd} effect -- the larger the dataset, the better protected a text becomes from its original. Finally, as dataset size grows, the typical DP utility hit becomes less apparent, and in some (\textit{mechanism}, $\varepsilon$) scenarios, one can achieve utility quite close to the non-private baselines (see AG News or Trustpilot, for example). This result is supported by the significance found in our regression model for increasing dataset size as a factor for higher relative gains. These findings shed light on the promise of DP text rewriting in practice and at scale.

However, there also come a few important considerations to this question. The first is a practical caveat: since the mechanisms tested in this work operate with \textit{local} DP, having larger dataset sizes would imply a data processor who is capable of (and trusted to) process larger data volumes, as it is unlikely that single users would possess such data. Secondly, it is still unclear what is the \textit{upper bound} of the effects observed in this work, and even larger-scale tests are required to continue investigating the effect of dataset size. For this, however, massive datasets would be needed, and furthermore, ones that can be reasonably attributed to some sensitive or adversarial scenario. This becomes crucial going forward, i.e., establishing high-quality and practical benchmarks for testing DP text privatization.

On this note, our findings point to an important consideration in the design of privacy benchmarking DP text rewriting. In our regression analysis, we find that the nature of both the utility and privacy tasks (as proxied by the number of associated labels) significantly impacts the quantification of relative gain. While further tests on a wider variety of downstream tasks would be needed to validate the regression coefficients, we nevertheless learn of the influence that evaluation setup decisions may have on the measurement (and eventual perception) of the effectiveness of DP text rewriting.

\section{Conclusion}
We investigate the effect of dataset size in DP text rewriting, using five datasets to evaluate both utility and privacy in various rewriting scenarios. Our results suggest that while larger dataset sizes are not a silver bullet for effective text privatization, they generally lead to more favorable trade-offs. This, however, is mechanism- and task-specific, showcasing the complexity of local DP text rewriting, and of DP NLP in general. Furthermore, we empirically demonstrate the importance of varying dataset size in evaluation procedures, showing that measured trade-offs may differ based on the size of the utilized datasets. In this, we provide an outlook for DP text privatization and evaluation at scale, proposing that with further (large-scale) testing, proposed techniques from the literature may begin to realize their potential in practice. 

\section*{Acknowledgments}
The authors would like to thank the reviewers for their helpful feedback in improving this work, as well as Alexandra Klymenko for her valuable support in preparing the paper.

\section*{Limitations}
A primary limitation with our work comes with the definition of \say{large-scale} datasets. While we utilized datasets that were known to us, publicly available, and reasonable to use within the confines of our resources, we are aware that the chosen datasets may still be considered small in the light of \say{big data}. However, we posit that the findings resulting from our experiments provide a starting point for even larger-scale experiments, especially in comparison to previous DP NLP works on smaller dataset samples.

Another limitation is one common to many related works: the evaluation of \textit{privacy}. We chose a combination of attribute-based empirical privacy (mitigation of inference attacks), as well as a designed indistinguishability test. We choose these in order to follow the evaluation procedure of previous works, especially in light of no standard benchmark for DP text rewriting. As such, the results we provide are a \textit{proxy} for privacy. Our relative gain ($\gamma$) calculations are therefore based on this proxy.

Finally, as showcased by our results, the effect of DP rewriting at scale can be significantly different from one mechanism to another. Our choice of inherently different rewriting mechanisms is a testament to that fact, yet we do not analyze in detail the potential technical origins of these differences, for example, in light of word- versus document-level rewriting. Future works could supplement our findings with such an analysis, particularly by including more than our chosen three mechanisms.

\section*{Ethics Statement}
As our work is focused in the field of privacy-preserving NLP, it places itself directly parallel to ethical and responsible NLP and AI. We hope that with the findings of our work, we may contribute to the body of knowledge of how to conduct (and evaluate) NLP in a privacy-preserving manner.

One ethical consideration comes with the simulation of adversaries in our privacy experiments. This is performed using publicly available datasets not originally intended for these adversarial purposes. Nevertheless, such considerations are mitigated since the datasets used (1) are long-standing in the research sphere, (2) contain no PII, and (3) have been used similarly by previous DP NLP works.

\bibliography{custom}

\appendix

\section{Relative Gain Calculation}
\label{sec:appendix_rg}
For the Trustpilot dataset, the complete dataset is heavily imbalanced towards positive reviews (93.07\%), yielding an overall $MG_u$ of 96.41\%. Accordingly, for the val set on all dataset splits, we calculate:
\begin{itemize}
    \itemsep 0em
    \item \textbf{10\%}: $MG_u$ = 96.36\%
    \item \textbf{25\%}: $MG_u$ = 96.38\%
    \item \textbf{50\%}: $MG_u$ = 96.47\%
    \item \textbf{75\%}: $MG_u$ = 96.54\%
    \item \textbf{100\%}: $MG_u$ = 96.41\%
\end{itemize}

Similarly, for the Yelp dataset, the reviews are primarily positive (78.74\%), for an overall $MG_u$ of 88.11\%. This yields the following values, calculated from the 10\% val split:
\begin{itemize}
    \itemsep 0em
    \item \textbf{10\%}: $MG_u$ = 90.05\%
    \item \textbf{25\%}: $MG_u$ = 88.60\%
    \item \textbf{50\%}: $MG_u$ = 88.30\%
    \item \textbf{75\%}: $MG_u$ = 88.66\%
    \item \textbf{100\%}: $MG_u$ = 88.69\%
\end{itemize}

Finally, for the Twitter dataset, the tweets are primarily positive (49.48\%), for an overall $MG_u$ of 66.20\%. This yields the following values, calculated from the 10\% val split:

\begin{itemize}
    \itemsep 0em
    \item \textbf{10\%}: $MG_u$ = 67.07\%
    \item \textbf{25\%}: $MG_u$ = 67.06\%
    \item \textbf{50\%}: $MG_u$ = 66.95\%
    \item \textbf{75\%}: $MG_u$ = 66.86\%
    \item \textbf{100\%}: $MG_u$ = 67.14\%
\end{itemize}

We also calculate the values for the smaller sample used for \textsc{DP-BART}:
\begin{itemize}
    \itemsep 0em
    \item \textbf{10\%}: $MG_u$ = 65.99\%
    \item \textbf{25\%}: $MG_u$ = 68.17\%
    \item \textbf{50\%}: $MG_u$ = 67.16\%
    \item \textbf{75\%}: $MG_u$ = 67.42\%
    \item \textbf{100\%}: $MG_u$ = 67.27\%
\end{itemize}

These values are used for their respective dataset splits in the $\gamma$ calculations for Tables \ref{tab:privacy} and \ref{tab:privacy_twitter}.

\end{document}